\documentclass[runningheads]{llncs}
\usepackage{algorithm, algorithmic}
\usepackage{amsmath,amssymb}
\usepackage{graphicx}
\usepackage{subfigure} 
\usepackage{multirow}
\usepackage{comment}
\usepackage{makecell}
\usepackage{rotating}
\usepackage{booktabs}
\usepackage{color}
\usepackage[misc]{ifsym}

\newcommand{\para}[1]{\smallskip\noindent\textbf{#1}}

\begin{document} 
\title{FAWA: Fast Adversarial Watermark Attack on Optical Character Recognition (OCR) Systems}

\titlerunning{FAWA: Fast Adversarial Watermark Attack on OCR Systems}

\author{Lu Chen\inst{1}, Jiao Sun\inst{2} \and Wei Xu\inst{1}{(\Letter)}}

\authorrunning{L. Chen et al.}

\institute{Institute for Interdisciplinary Information Sciences, Tsinghua University, Beijing 100084, China\\
\email{lchen17@mails.tsinghua.edu.cn, weixu@mail.tsinghua.edu.cn} \and
CS, University of Southern California, Los Angeles, CA 90007, USA\\
\email{jiaosun@usc.edu}}

\maketitle   \setcounter{footnote}{0}           

\begin{abstract}
Deep neural networks (DNNs) significantly improved the accuracy of optical character recognition (OCR) and inspired many important applications.
Unfortunately, OCRs also inherit the vulnerabilities of DNNs under adversarial examples.  
Different from colorful vanilla images, text images usually have clear backgrounds. 
Adversarial examples generated by most existing adversarial attacks are unnatural and pollute the background severely. 
To address this issue, we propose the  \emph{\textbf{F}ast \textbf{A}dversarial \textbf{W}atermark \textbf{A}ttack~(FAWA)} against sequence-based OCR models in the white-box manner.  
By disguising the perturbations as watermarks, we can make the resulting adversarial images appear natural to human eyes and achieve a perfect attack success rate.  
FAWA works with either gradient-based or optimization-based perturbation generation.  
In both letter-level and word-level attacks, our experiments show that in addition to natural appearance, FAWA achieves a 100\% attack success rate with 60\% less perturbations and 78\% fewer iterations on average.  
In addition, we further extend FAWA to support full-color watermarks, other languages, and even the OCR accuracy-enhancing mechanism. 

\end{abstract}

\section{Introduction}

Optical Character Recognition (OCR) has been an important component in text processing applications, such as license-plate recognition, street sign recognition and financial data analysis.  Deep neural networks (DNNs) significantly improve OCR's performance.  
Unfortunately, OCR inherits all counter-intuitive security problems of DNNs. OCR models are also vulnerable to \emph{adversarial examples}, which are crafted by making human-imperceptible perturbations on original images to mislead the models. 
The wide application of OCR provides incentives for adversaries to attack OCR models, thus damaging downstream applications, resulting in fake ID information, incorrect metrics readings and  financial data. 





Prior works have shown that we can change the prediction of DNNs in image classification tasks only by applying carefully-designed perturbations~\cite{goodfellow2014explaining,nguyen2015deep,papernot2016limitations,szegedy2013intriguing}, or adding a small patch~\cite{brown2017adversarial,karmon2018lavan} to original images.  
But these methods are not directly applicable to OCR attacks. 
1) Most text images are on white backgrounds. Perturbations added by existing methods appear too evident for human eyes to hide, and pollute clean backgrounds. 
2) Instead of classifying characters individually, modern OCR models are segmentation-free, inputting entire variable-sized image and outputting a sequence of labels. It is called the \textit{sequential labeling} task~\cite{graves2006connectionist}. 
Since modern OCR models use the CNN~\cite{krizhevsky2012imagenet} and the LSTM~\cite{hochreiter1997long} as feature extractors, the internal feature representation also relies on contexts (i.e., nearby characters).  
Therefore, besides perturbing the target character, we also need to perturb its context, resulting in more perturbations. 
3) Since text images are usually dense, there is no open space to add a patch.  




In this paper, we propose a new attack method, \textit{Fast Adversarial Watermark Attack}~(FAWA), against modern OCR models. 
Watermarks are images or patterns commonly used in documents as  backgrounds to identify things, such as marking a document proprietary, urgent, or merely for decoration.   
Human readers are so used to the watermark that they naturally ignore it. 
We generate natural watermark-style adversarial perturbations in text images. Such images appear natural to human eyes but misguide  OCR models. 
Watermark perturbations are similar to patch-based attacks~\cite{brown2017adversarial,karmon2018lavan} where perturbations are confined to a region. 
Different from patches occupying a separate region, watermarks overlay on texts but not hinder the text readability. 
Laplace attack~\cite{hanwei2019smooth} and HAAM~\cite{heng2018harmonic} try to generate seemingly smooth perturbations, which work well on colored photos. 
However, they do not solve the problem of background pollution.


FAWA is a \emph{white-box} and \emph{targeted} attack. 
We assume adversaries have perfect knowledge of the DNN architecture and parameters (white-box model) and generate specific recognition results (targeted).  
There are three steps in the perturbation generation. 
1) We find a good position over the target character to add an initial watermark. 
2) We generate perturbations inside the watermark. 
3) Optionally, we convert the gray watermark into a colored one.  
To generate perturbations, we leverage either gradient-based or optimization-based methods.  
We evaluate FAWA on a state-of-the-art open-source OCR model, Calamari-OCR~\cite{wick2018calamari} for English texts with five fonts. 
FAWA generates adversarial images with 60\% less perturbations and 78\% fewer iterations than existing methods, while maintaining a 100\% attack success rate.  
Our adversarial images also look quite similar to natural watermarked images.  
We evaluate the effects of colored watermarks and other languages under real-world application settings.
Last, we propose a positive application of the FAWA, i.e., using the perturbations to enhance the accuracy of OCR models.  
The contributions of this paper include:
\begin{enumerate}
	\item We propose an attack method, FAWA,  of hiding adversarial perturbations in watermarks from human eyes. We implemented FAWA as the efficient adversarial attacks against the DNN-based OCR in sequential labeling tasks;
	\item Extensive experiments show that FAWA performs targeted attacks perfectly, and generate natural watermarked images with imperceptible perturbations;
	\item We demonstrate several applications of FAWA, such as colored watermarks as an attack mechanism, using FAWA as an accuracy-enhancing mechanism.
\end{enumerate}

\section{Background and Related Work}

\para{Optical Character Recognition (OCR).  }
We can roughly categorize existing OCR models into character-based models and end-to-end models.
\emph{Character-based recognition models} segment the image into character images first, before passing them into the recognition engine. Obviously, the OCR performance heavily relies on the character segmentation. 
\emph{End-to-end recognition models} apply an unsegmented recognition engine, which recognizes the entire sequence of characters in a variable-sized image. \cite{bengio1995lerec,espana2011improving} adopt sequential models such as Markov models and \cite{breuel2013high,wang2012end} utilize DNNs as feature extractors for sequential recognition. \cite{graves2006connectionist} introduces a segmentation-free approach, connectionist temporal classification~(CTC), which provides a sort of loss function of enabling us to recognize variable-length sequences without explicit segmentation while training DNN models. 
Thus, many state-of-the-art OCR models use CTC as the loss function.

\para{Attacking DNN-based Computer Vision Models.  }
Attacking DNN-based models is a popular topic in both computer vision and security fields.  
Existing attack methods use the following two ways to generate perturbations. 
1) Making perturbations small enough for evading human perception.  
For example, many projects aim at generating minor $\mathrm{L_p}$-norm perturbations for the purpose of making them visually imperceptible. 
FGSM~\cite{goodfellow2014explaining}, L-BFGS~\cite{szegedy2013intriguing}, DeepFool~\cite{moosavi2016deepfool}, C\&W~{$\mathrm{L_2}/ \mathrm{L_{\infty}}$}~\cite{carlini2017towards}, PGD~\cite{madry2017towards} and EAD~\cite{chen2018ead}, all perform modifications at the pixel level by a small amount bounded by $\epsilon$.  
2) Making perturbations in a small region of an image.  
For example, JSMA~\cite{papernot2016limitations}, C\&W $\mathrm{L_0}$~\cite{carlini2017towards}, Adversarial Patch~\cite{brown2017adversarial} and LaVAN~\cite{karmon2018lavan}, all perturb a small region of pixels in an image, but their perturbations are not limited by $\epsilon$ at the pixel level. 
Though FAWA is similar to patches, in the text images, watermarks overlay on the text instead of covering the text.  



\para{Generating minimal adversarial perturbations.}  
FAWA generates adversarial perturbations using the either gradient-based or optimization-based methods.

Gradient-based methods add perturbations generated from gradient against input pixels. 
\emph{FGSM}~\cite{goodfellow2014explaining} is a $\mathrm{L_{\infty}}$-norm one-step attack.
It is efficient but only provides a coarse approximation of the optimal perturbations. 
\emph{BIM}~\cite{kurakin2016adversarial} takes multiple smaller steps and the resulting image is clipped by the same bound $\epsilon$. Thus BIM produces superior results to FGSM. 
\emph{MI-FGSM}~\cite{dong2018boosting} extends BIM with momentum. MI-FGSM can not only stabilize update directions but also escape from poor local maxima during iterations, and thus generates more transferable adversarial examples. Considering the efficiency, we adopt MI-FGSM in FAWA. 

Optimization-based methods directly solve the box-constrained optimization problem to minimize the $\mathrm{L_p}$-norm distance between the original and adversarial image, while yielding the targeted result.  
\emph{Box-constraint L-BFGS}~\cite{szegedy2013intriguing} seeks adversarial examples with L-BFGS. Though L-BFGS constructs subtle perturbations, it is inefficient. 
Instead of applying cross-entropy as loss function, \emph{C\&W ${L_2}$ attack}~\cite{carlini2017towards} constructs a new loss function and solves it with gradient descent. 
\emph{OCR attack}~\cite{Song2018Fooling} is the only available work of attacking sequence-based OCRs as far as we know, which generates adversarial examples using the CTC loss for sequential labeling tasks.  
In FAWA, we use the same setting as OCR attack.



\para{Perturbations with other optimization goals.} 
Besides minimizing the perturbation level, many works make efforts to produce smooth and natural perturbations. 
Laplace attack~\cite{hanwei2019smooth} smooths perturbations relying on Laplacian smoothing. 
HAAM~\cite{heng2018harmonic} creates edge-free perturbations using harmonic functions for disguising natural shadows or lighting. 
However, the smoothing and disguising are for photos but not for text images. 
Instead of manipulating pixel values directly, \cite{Xiao2018SpatiallyTA} produces perceptually realistic perturbations with spatial transformation. Though avoiding background pollution, it cannot guarantee the readability of text when the attack needs large deformation.
\cite{wang2019adversarial} also utilizes the watermark idea but performs attacks only by scaling and rotating plain watermarks. Without adding pixel-level perturbations, it does not offer a high attack success rate.



\section{Fast Adversarial Watermark Attack}
In this section, we introduce \textit{fast adversarial watermark attacks}~(FAWA). 
FAWA consists of three steps. 
1) We automatically determine a good position to add the initial watermark in the text image so that we can confine the perturbation generation to that region.
2) We generate the watermark-style perturbations with either the \textit{gradient-based method} or \textit{optimization-based method}.
3) Optionally, we convert gray watermarks into full-color ones to improve the text readability.

\subsection{Preliminaries}
\para{Problem definition of adversarial image generation. }
Given a text image $\boldsymbol{x}=[x_1,x_2, ..., x_n]^\mathrm{T} $ for any $x_i \in  [0, 1]$, where $n$ is the number of pixels in the image, our goal is to generate an adversarial example $\boldsymbol{x}'$ with minimum $\mathrm{L_p}$-norm perturbations $\|\boldsymbol{x}^{\prime}-\boldsymbol{x}\|_{p}$ against the white-box model $f$ with intent to trick model $f$ into outputting the targeted result $\boldsymbol{t}$, $f(\boldsymbol{x'})=\boldsymbol{t}$. 
Formally, the problem of adversarial image generation is $
	\min _{\boldsymbol{x}^{\prime}}~ \|\boldsymbol{x}^{\prime}-\boldsymbol{x}\|_{p} ~ \text { s.t. }  f(\boldsymbol{x'})=\boldsymbol{t}~\text{and}~ \boldsymbol{x}^{\prime} \in[0,1]^{n}$.
Also, we define the \emph{CTC loss function} with respect to the image $\boldsymbol{x}$ as $\ell_{\text{CTC}}(\boldsymbol{x}, \boldsymbol{t})$  and the target labels $\boldsymbol{t} = [t_1, t_2, ..., t_N]$ for $t_i \in \mathcal{T}$, where $\mathcal{T}$ is the character set.

\para{Saliency map.} 
The saliency map~\cite{papernot2016limitations} is a versatile tool that not only provides us valuable information to cause the targeted misclassification in the threat model but also assists us in intuitively explaining some attack behaviors as a visualization tool. 
The saliency map indicates the output sensitivity, relevant to the adversarial targets, to the input features.
In the saliency map, a larger value indicates a higher likelihood of fooling the model $f$ to output the target $\boldsymbol{t}$ instead of the ground-truth $f(\boldsymbol{x})$.
We can construct a saliency map of an image using the forward derivative with respect to each input component in the text image. 

\para{$\mathrm{L_p}$-norms.}  
$\mathrm{L_p}$-norms are commonly-used metrics to measure the perceptual similarity between the clean image $\boldsymbol{x}$ and the adversarial image $\boldsymbol{x}'$, denoted by 
$\|\boldsymbol{x} - \boldsymbol{x}'\|_p =(\sum_{i=i}^{n} |\boldsymbol{x} - \boldsymbol{x}'_i|^p)^{\frac{1}{p}}$, $p=2, \infty$. 
In gradient-based methods, we apply $\mathrm{L_p}$-norms to prune the saliency map for generating $\mathrm{L_p}$-norm perturbations. 
In optimization-based methods, $\mathrm{L_p}$-norms are usually as an optimization term in the objective function. 
Particularly, \textit{${L_2}$-norm} is greatly useful to enhance the visual quality.
 \textit{$L_{\infty}$-norm} is to measure the maximum variation of perturbations.

\subsection{Finding the Position of Watermarks} 
\label{sec:wm_pos}
\begin{figure}[tb]
\centering
\includegraphics[width=\linewidth]{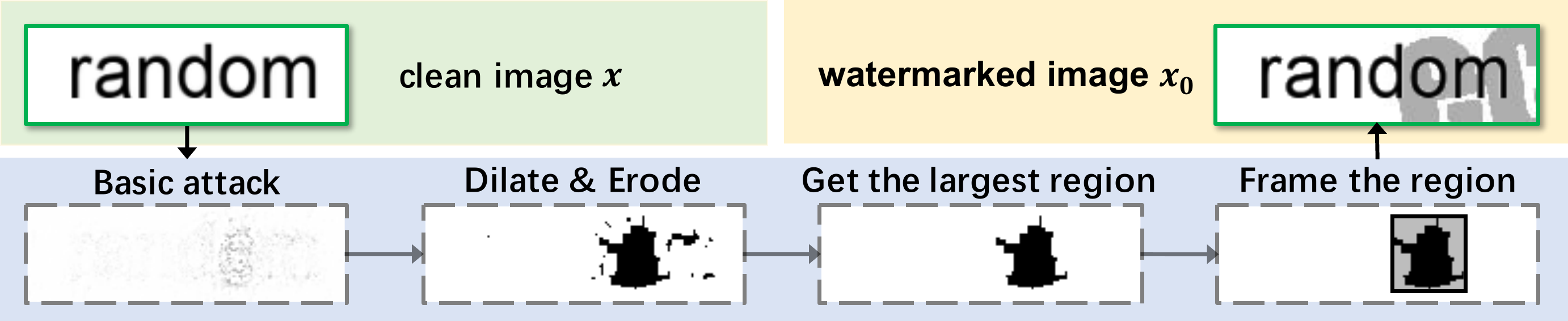}
\caption{Find the position of watermarks.}
\label{fig:find_pos}
\end{figure}

To automatically determine a good position of watermarks, as shown in Fig. \ref{fig:find_pos}, we perform the following steps.
 1) We produce adversarial perturbations of the basic attack (i.e., Grad-Basic or Opt-Basic in Section~\ref{sec:wm_attack}) in the text image. 
 2) We binarize such adversarial image, and get its perturbed regions $\boldsymbol{r}=[r_1, r_2, ..., r_n]^\mathrm{T}$, where $r_i=1$ if $x_i \neq x'_i$ and 0 otherwise. 
 3) In order to find relatively complete perturbed regions $\boldsymbol{r}'$, we apply a combination of erosion $\oplus$ and dilation $\ominus$ operations twice in the perturbed regions $\boldsymbol{r}$, $\boldsymbol{r}'=((\boldsymbol{r} \oplus \mathcal{K}) \ominus \mathcal{K})^2$, where we set the kernel $\mathcal{K}=\boldsymbol{1}_{3 \times 3}$. 
 4) After sorting the perturbed regions in $\boldsymbol{r}'$ by their area, we obtain the largest perturbed region in $\boldsymbol{r}'$. We can find that our target texts locate in the same position as the largest found region.
 5) Last, in the text image, we place a watermark big enough to cover the found position, and obtain an initial watermarked image $\boldsymbol{x}_0$ and a binary watermark mask $\Omega_w$ with the same shape of $\boldsymbol{x}$, where $\Omega_{w,i}=1$ if the position i is inside the watermark and 0 otherwise. 
 
%

\subsection{Generating Adversarial Watermarks}
\label{sec:wm_attack}
After finding the position of watermarks, we need to generate the perturbations within the watermarks to mislead the OCR models to output target texts. 
Integrated with two popular methods, we use the following methods to attack. 

\para{Gradient-based Watermark Attack~(Grad-WM).}
\label{sec:wm_grad}
Considering the high efficiency of MI-FGSM~\cite{dong2018boosting}, we apply it  as our gradient-based method (we refer to as \emph{Grad-Basic}). 
Each iterative update of the Grad-Basic is to 1) get the saliency map normalized by $\mathrm{L}_1$-norm with the cross-entropy loss, 2) adjust the update direction in the saliency map and update the momentum, and 3) update a new $\mathrm{L_p}$-norm adversarial image $\boldsymbol{x}'_{i+1}$ bounded by $\epsilon$ using the updated momentum.

There are two main differences between Grad-WM and Grad-Basic.
1) Different from off-the-shelf MI-FGSM, where it applies the cross-entropy as loss function in the image classification tasks, in our Grad-WM, we use the CTC loss function to compute the saliency map. CTC loss function fits better because it is widely used in OCR models to handle the sequential labeling tasks. 
2) In Grad-WM, to hide perturbations in the watermarks, we confine the perturbations within the boundary of the watermark $\Omega_w$, rather than spreading perturbations over the entire image in the result of the background pollution like Grad-Basic.


Algorithm \ref{alg:1} summarizes the Grad-WM generation procedure. 
Notably, input images are the initial watermarked images $\boldsymbol{x}_0$ created in Section~\ref{sec:wm_pos}. Then the saliency map is produced with the CTC loss function relevant to the adversarial targets $\boldsymbol{t}$. Through element-wisely multiplying the watermark mask $\Omega_w$ with the updated $\mathrm{L_p}$-norm $\boldsymbol{g}_{i+1}$, we get rid of the perturbations outside the watermark to maintain the clean background. We gain a significant visual improvement in the Grad-WM than the Grad-Basic. 
 Besides, for further improving the attack efficiency, we determine whether to stop the attacks in every few iterations. 

\begin{algorithm}[tb]
    \footnotesize
    \renewcommand{\algorithmicrequire}{\textbf{Input:}}
    \renewcommand{\algorithmicensure}{\textbf{Output:}}
    \caption{Gradient-based Watermark Attack}
    \label{alg:1}
    \begin{algorithmic}[1]
        \REQUIRE A text image $\boldsymbol{x}$, OCR model $f$ with CTC loss $\ell_{\text{CTC}}$, targeted text $\boldsymbol{t}$, 
        $\epsilon$-bounded perturbation, attack step size $\alpha$, \# of maximum iterations $I$, decay factor $\mu$.
        \ENSURE An adversarial example $\boldsymbol{x}'$ with $\|\boldsymbol{x}' - \boldsymbol{x}\|_p \leq \epsilon$ or attack failure $\bot$.
        \STATE Initialization: $\boldsymbol{g}_0 = \boldsymbol{0}$; $\boldsymbol{x}_0' = \boldsymbol{x}_0$
        \FORALL{each iteration $i = 0$ to $I-1$}
        \STATE Input $\boldsymbol{x}'_{i}$ to $f$ and obtain the saliency map $\nabla_{\boldsymbol{x}} \ell_{\text{CTC}}(\boldsymbol{x}'_i, \boldsymbol{t})$
        \STATE Update $\boldsymbol{g}_{i+1}$ by accumulating $\boldsymbol{g}_i$ in the saliency map normalized by  $\mathrm{L_1}$-norm as 
        \begin{align} 
        \label{eq:momentum}
        \boldsymbol{g}_{i+1}=\mu \cdot \boldsymbol{g}_{i}+\frac{\nabla_{\boldsymbol{x}} \ell_{\text{CTC}}(\boldsymbol{x}_{i}', \boldsymbol{t})}{\|\nabla_{\boldsymbol{x}} \ell_{\text{CTC}}(\boldsymbol{x}_{i}', \boldsymbol{t})\|_{1}} 
        \end{align}
        \STATE Update $\boldsymbol{x}_{i+1}'$ by applying watermark-bounded $\mathrm{L_p}$-norm perturbations as 
        \begin{align} 
        \label{fmt:adv_x}
        \boldsymbol{x}_{i+1}' = \boldsymbol{x}_{i}' + \text{clip}_{\epsilon}(\alpha \cdot (\Omega_w \odot \frac{\boldsymbol{g}_{i+1}}{\|\boldsymbol{g}_{i+1}\|_p}))
        \end{align}
        \IF{$f(\boldsymbol{x}'_{i+1}) == \boldsymbol{t}$}
        \RETURN $\boldsymbol{x}'_{i+1}$
        \ENDIF
        \ENDFOR
        \RETURN attack failure $\bot$
\end{algorithmic}
\end{algorithm}

\para{Optimization-based Watermark Attack~(Opt-WM).}
We employ OCR attack~\cite{Song2018Fooling} as our basic optimization-based method~(Opt-Basic).  Opt-Basic solves the following optimization problem: $\min _{\boldsymbol{x}'} ~ c \cdot \ell_{\text{CTC}}(\boldsymbol{x}',$ $\boldsymbol{t})$ + $\|\boldsymbol{x}'-\boldsymbol{x}\|_2^2$ s.t. $\boldsymbol{x}'\in [0, 1]^n$, where $c$ is a hyper-parameter. 
To improve the visual quality of adversarial images, Opt-Basic adopts $\mathrm{L_2}$-norm for penalizing perturbations. 
To eliminate the box-constraint $[0, 1]$ of $\boldsymbol{x}'$, it reformulates the problem using the change of variables~\cite{carlini2017towards} as $\min _{\boldsymbol{\omega}} ~ c \cdot \ell_{\text{CTC}}(\frac{ \tanh (\boldsymbol{\omega})+1}{2}, \boldsymbol{t})$ + $\|\frac{\tanh (\boldsymbol{\omega})+1}{2}-\boldsymbol{x}\|_2^2$, which optimizes a new variable $\boldsymbol{\omega}$ instead of the box-constrained $\boldsymbol{x}'$. 
The fact $-1 \leq \tanh (\cdot) \leq 1$ implies that $\frac{ \tanh (\cdot)+1}{2}$ satisfies the box-constraint $[0, 1]$ automatically. 
Intuitively, we treat the variable $\boldsymbol{\omega}$ as the adversarial example $\boldsymbol{x}'$ without the box-constraint.

Similar to Grad-WM, we first get the initial watermarked input image $\boldsymbol{x}_0$ and the watermark mask $\Omega_w$. 
To confine perturbations in watermarks more conveniently, we separate the perturbation term from $\boldsymbol{w}$, and rewrite the original $\boldsymbol{w}$ to $\boldsymbol{w} + \boldsymbol{x}_0$ that represents the adversarial image of combining the perturbation term $\boldsymbol{w}$ and the initial watermarked image term $\boldsymbol{x}_0$. The objective function changes to $\min _{\boldsymbol{\omega}} ~ c \cdot \ell_{\text{CTC}}(\frac{ \tanh (\boldsymbol{\omega}+\boldsymbol{x}_0)+1}{2}, \boldsymbol{t}) + \|\frac{\tanh (\boldsymbol{\omega}+\boldsymbol{x}_0)+1}{2}-\boldsymbol{x}_0\|_2^2$.
To constrain the manipulation region in the watermark fashion, we perform the element-wisely multiplication between the perturbation variable $\boldsymbol{\omega}$ and the watermark mask $\Omega_w$. 
Formally, we reformulate the optimization problem as 
\begin{align}
\min _{\boldsymbol{\omega}} ~ c \cdot \ell_{\text{CTC}}(\frac{ \tanh (\Omega_w \odot \boldsymbol{\omega}+\boldsymbol{x})+1}{2}, \boldsymbol{t}) + \|\frac{\tanh (\Omega_w \odot \boldsymbol{\omega}+\boldsymbol{x})+1}{2}-\boldsymbol{x}\|_2^2.
\label{eq:opt_wm}
\end{align}
We adopt the Adam optimizer~\cite{kingma2014adam} for both Opt-Basic and Opt-WM to seek the watermark-style adversarial images. 
 We utilize the binary search to adapt the tradeoff hyper-parameter $c$ between the loss function and the $\mathrm{L_2}$-norm distance.



 \subsection{Improving Efficiency} 
 \label{sec:improve_efficiency}
  To improve attack efficiency, we employ batch attack and early stopping by increasing the number of parallel attacks and reducing redundant attack iterations.

\para{Batch attack.} 
Because OCR models only support batch image processing with the same size, 
when we attack a large number of variable-size text images, it will be time-consuming to attack them one by one. 
To improve attack parallelism (i.e., attack multiple images simultaneously), we 1) resize the variable-size images into the same height, 2) pad them into the maximum width among these images, 3) put the same-size images into a matrix. 
After the preprocessing, we could perform batch adversarial attacks in a single matrix to improve attack efficiency. 

\para{Early stopping.}
For avoiding redundant attack iterations, we insert the early-stop mechanism in the attack iterations. 
We evaluate attack success rate~(ASR) every few iterations.
We stop attack iteration once ASR achieves 100\%, indicating that all adversarial images are misidentified as the targeted texts by the threat model. 
To a certain extent, early stopping reduces the attack efficiency as it takes time to check attack status during iterations.  
This is a basic tradeoff between the cycle of ASR evaluation and the maximum attack iteration setting.

\subsection{Improving Readability} 
Sometimes the gray watermarks still hinder the readability of the text as it reduces the contrast.  To achieve better readability of the text, we employ the following two effective strategies: adding text masks and full-color watermark.

\para{Adding Text Masks}. 
When generating the initial watermarked image $\boldsymbol{x}_0$ in Section~\ref{sec:wm_pos}, to prevent the text from being obscured by the watermarks, we add in watermarks outside the text. 
To achieve this, we binarize the text images $\boldsymbol{x}$ by a threshold $\tau$ to get text masks $\Omega_t$ with the same size of $\boldsymbol{x}$, where $\Omega_{t, i} =1$ if $x_i > \tau$ and 0 otherwise. 
Then we get the initial watermarked images $\boldsymbol{x}_0 =   \boldsymbol{x} \odot (\mathbf{1} - \Omega_w \odot \overline{\Omega_t}) + \beta \cdot \Omega_w \odot \overline{\Omega_t} $, where $\beta$ is the grayscale value of initial watermarks.

\para{Full-Color Watermark}. 
Modern OCR systems always preprocess colored images into grayscale ones before recognition. 
Compared to grayscale watermarks, colored watermarks have better visual quality on the black texts. 
To convert grayscale watermarks into RGB ones, we only manipulate the pixel $x_i$ inside the watermark, where $\Omega_{w, i}=1$. 
Given a grayscale value~\textit{Gray} in the grayscale watermark, we preset \textit{R} value and \textit{B} value to certain values. 
Next we make the color conversion with the transform equation: ${Gray} = {R} * 0.299 + {G} * 0.587 + {B} * 0.114$, to calculate the left \textit{G} value.
Last we get RGB values from a grayscale value. 



\section{Experiments}

\subsection{Experiment Setup}

\para{OCR model.}  We choose to attack Calamari-OCR \footnote{https://github.com/Calamari-OCR/calamari} \cite{wick2018calamari}. The OCR model has two convolutional layers, two pooling layers, and the following LSTM layer. We use the off-the-shelf English model as our threat model, trained with CTC loss.

\para{Generate text images.  } 
After processing the corpus in the IMDB dataset, we generate a dataset with 97 paragraph, 1479 sentence and 1092 word images using the Text Recognition Data Generator\footnote{https://github.com/Belval/TextRecognitionDataGenerator}.
We use five fonts to generate these images: Courier, Georgia, Helvetica, Times and Arial.  
We set the font size to 32 pixels. 
We verify the Calamari-OCR can achieve 100\% accuracy on these images.


 
%

\para{Generating letter-level attacks targets.}
We choose a target word that is 1) a valid word, and 2) with edit distance 1 from the original.  
We evaluate replacement, insertion and deletion.  
More similar the replacement target is to the original, the easier the attack is.  
For example, replacing letter \texttt{t} with letter \texttt{f} is easier than replacing \texttt{t} with letter \texttt{j}.  
We use the \textit{logit} value from the output of the last hidden layer, as the similarity metric between a pair of letters.  
Given the logit value, we assign replacement attacks into easy, random and hard case.

\para{Generating word-level attacks targets.}
In this task, we replace the entire word in word, sentence and paragraph images.  
Different from letter-level attacks, we randomly choose a word in the English dictionary which has the same length as the original. However, we don't constrain the edit distance between them.



\para{Attack implementation and settings.}
We implemented Grad-WM and Opt-WM attacks\footnote{https://github.com/strongman1995/Fast-Adversarial-Watermark-Attack-on-OCR}. 
We normalize all input images to $[0, 1]$ and set $\beta$, the grayscale value of initial watermarks, to 0.682.
Given color transform equation, fixing $R=255$ and $B=0$, RGB's upper bound (255, 255, 0) is equal to $Gray=0.882$. After adding $\epsilon$-bounded perturbations ($\epsilon=0.2$), watermarks are still less than 0.882, the upper bound for valid conversion.
We use the watermark style: the word ``ecml" of Impact font with 78-pixel font size and 15-degree rotation.  

For the gradient-based methods, we use the implementation of MI-FGSM in CleverHans Python library\footnote{https://github.com/tensorflow/cleverhan}. 
We set maximum iterations $I=2000$, $\text{batch size}=100$ and $\epsilon=0.2$. 
For $\mathrm{L_2}$-norm, $\alpha=0.05$; for $\mathrm{L_{\infty}}$-norm, $\alpha=0.01$. 
The momentum decay factor $\mu$ is 1.0.
For the optimization-based methods, we use the Adam optimizer~\cite{kingma2014adam} for 1000 steps with mini-batch size 100 and learning rate 0.01.
 We choose $c=10$ as the tradeoff between the targeted loss and perturbation level.  
 


\begin{table}[tb]
\begin{center}
\resizebox{\textwidth}{!}{
\begin{tabular}{| c | c | c | c | c | c | c | c | c |} 
\hline
 & \scriptsize{original $\boldsymbol{x}$}
 & \scriptsize{$\text{WM}_0$ $\boldsymbol{x}_0'$}
 & \scriptsize{Grad-Basic}
 & \scriptsize{Grad-WM}
 & \scriptsize{Opt-Basic} 
 & \scriptsize{Opt-WM}
 & \scriptsize{Color-WM}
 & \scriptsize{target}\\ \hline

  \multirow{3}{*}{\rotatebox{90}{ letter ~} } &
 \begin{minipage}[]{0.11\textwidth}
 \centerline{ \includegraphics[width=\linewidth]{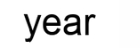}}
  \end{minipage} & 
  \begin{minipage}[]{0.11\textwidth}
   \centerline{\includegraphics[width=\linewidth]{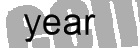}}
    \end{minipage}  & 
   \begin{minipage}[]{0.11\textwidth}
 \centerline{ \includegraphics[width=\linewidth]{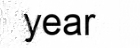}}
  \end{minipage} & 
  \begin{minipage}[]{0.11\textwidth}
 \centerline{ \includegraphics[width=\linewidth]{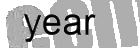}}
  \end{minipage} & 
  \begin{minipage}[]{0.11\textwidth}
 \centerline{ \includegraphics[width=\linewidth]{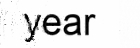}}
  \end{minipage} & 
  \begin{minipage}[]{0.11\textwidth}
 \centerline{ \includegraphics[width=\linewidth]{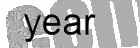}}
  \end{minipage} &
  \begin{minipage}[]{0.11\textwidth}
 \centerline{ \includegraphics[width=\linewidth]{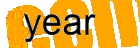}}
  \end{minipage} &
   hear
  \\  
  
 &
  \begin{minipage}[]{0.11\textwidth}
 \centerline{ \includegraphics[width=\linewidth]{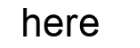}}
  \end{minipage} & 
  \begin{minipage}[]{0.11\textwidth}
   \centerline{\includegraphics[width=\linewidth]{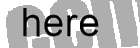}}
    \end{minipage}  & 
   \begin{minipage}[]{0.11\textwidth}
 \centerline{ \includegraphics[width=\linewidth]{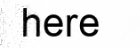}}
  \end{minipage} & 
  \begin{minipage}[]{0.11\textwidth}
 \centerline{ \includegraphics[width=\linewidth]{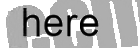}}
  \end{minipage} & 
  \begin{minipage}[]{0.11\textwidth}
 \centerline{ \includegraphics[width=\linewidth]{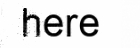}}
  \end{minipage} & 
  \begin{minipage}[]{0.11\textwidth}
 \centerline{ \includegraphics[width=\linewidth]{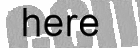}}
  \end{minipage} &
  \begin{minipage}[]{0.11\textwidth}
 \centerline{ \includegraphics[width=\linewidth]{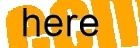}}
  \end{minipage} &
   there
  \\  
  
  &
  \begin{minipage}[]{0.11\textwidth}
  \centerline{ \includegraphics[width=\linewidth]{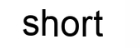}}
  \end{minipage} & 
  \begin{minipage}[]{0.11\textwidth}
   \centerline{\includegraphics[width=\linewidth]{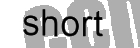}}
    \end{minipage}  & 
   \begin{minipage}[]{0.11\textwidth}
 \centerline{ \includegraphics[width=\linewidth]{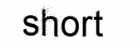}}
  \end{minipage} & 
  \begin{minipage}[]{0.11\textwidth}
 \centerline{ \includegraphics[width=\linewidth]{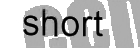}}
  \end{minipage} & 
  \begin{minipage}[]{0.11\textwidth}
 \centerline{ \includegraphics[width=\linewidth]{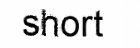}}
  \end{minipage} & 
  \begin{minipage}[]{0.11\textwidth}
 \centerline{ \includegraphics[width=\linewidth]{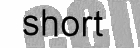}}
  \end{minipage} &
  \begin{minipage}[]{0.11\textwidth}
 \centerline{ \includegraphics[width=\linewidth]{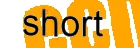}}
  \end{minipage} &
   sort
  \\    \hline

 \multirow{3}{*}{\rotatebox{90}{ \makecell{ word \qquad } }} &
 \begin{minipage}[]{0.11\textwidth}
 \centerline{ \includegraphics[width=\linewidth]{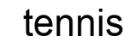}}
 \end{minipage} &
  \begin{minipage}[]{0.11\textwidth}
   \centerline{\includegraphics[width=\linewidth]{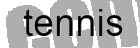}}
    \end{minipage}  & 
   \begin{minipage}[]{0.11\textwidth}
 \centerline{ \includegraphics[width=\linewidth]{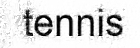}}
  \end{minipage} & 
  \begin{minipage}[]{0.11\textwidth}
 \centerline{ \includegraphics[width=\linewidth]{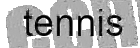}}
  \end{minipage} & 
  \begin{minipage}[]{0.11\textwidth}
 \centerline{ \includegraphics[width=\linewidth]{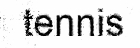}}
  \end{minipage} & 
  \begin{minipage}[]{0.11\textwidth}
 \centerline{ \includegraphics[width=\linewidth]{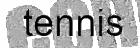}}
  \end{minipage} &
  \begin{minipage}[]{0.11\textwidth}
 \centerline{ \includegraphics[width=\linewidth]{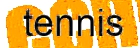}}
  \end{minipage} &
   amazon
  \\  

 &
\begin{minipage}[]{0.11\textwidth}
  \centerline{ \includegraphics[width=\linewidth]{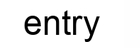}}
  \end{minipage} & 
  \begin{minipage}[]{0.11\textwidth}
   \centerline{\includegraphics[width=\linewidth]{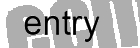}}
    \end{minipage}  & 
   \begin{minipage}[]{0.11\textwidth}
 \centerline{ \includegraphics[width=\linewidth]{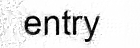}}
  \end{minipage} & 
  \begin{minipage}[]{0.11\textwidth}
 \centerline{ \includegraphics[width=\linewidth]{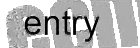}}
  \end{minipage} & 
  \begin{minipage}[]{0.11\textwidth}
 \centerline{ \includegraphics[width=\linewidth]{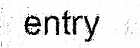}}
  \end{minipage} & 
  \begin{minipage}[]{0.11\textwidth}
 \centerline{ \includegraphics[width=\linewidth]{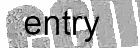}}
  \end{minipage} &
  \begin{minipage}[]{0.11\textwidth}
 \centerline{ \includegraphics[width=\linewidth]{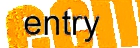}}
  \end{minipage} &
   waken
  \\   
   
   &
\begin{minipage}[]{0.11\textwidth}
  \centerline{ \includegraphics[width=\linewidth]{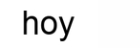}}
  \end{minipage} & 
  \begin{minipage}[]{0.11\textwidth}
   \centerline{\includegraphics[width=\linewidth]{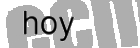}}
    \end{minipage}  & 
   \begin{minipage}[]{0.11\textwidth}
 \centerline{ \includegraphics[width=\linewidth]{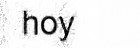}}
  \end{minipage} & 
  \begin{minipage}[]{0.11\textwidth}
 \centerline{ \includegraphics[width=\linewidth]{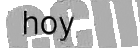}}
  \end{minipage} & 
  \begin{minipage}[]{0.11\textwidth}
 \centerline{ \includegraphics[width=\linewidth]{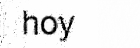}}
  \end{minipage} & 
  \begin{minipage}[]{0.11\textwidth}
 \centerline{ \includegraphics[width=\linewidth]{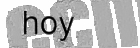}}
  \end{minipage} &
  \begin{minipage}[]{0.11\textwidth}
 \centerline{ \includegraphics[width=\linewidth]{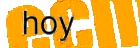}}
  \end{minipage} &
   jow
  \\   \hline
\end{tabular}}
\end{center}
\caption{Adversarial examples of letter-level and word-level attacks. Colored watermark examples are converted from Grad-WM examples.}
\label{tab:adv_example}
\end{table}

\para{Evaluation metrics. }
We evaluate attack capability from the following aspects.
\begin{description}
	\item[1) Perturbation level.] We quantify the perturbation level with three metrics:  
	\textit{MSE} $=\frac{1}{n} \|\boldsymbol{x} - \boldsymbol{x}'\|_2^2$, evaluates the difference between two images. 
	\textit{PSNR} $=10 \log \left(\frac{D^{2}}{\mathrm{MSE}}\right)$, where $D$ is the range of pixel intensities.
	\textit{SSIM}~\cite{wang2004image} captures structural information  and measures the similarity between two images. 
	For these metrics, we take the average of all images in the following results. 
	Smaller MSE, larger PSNR and SSIM closer to 1 indicate less perturbations.
	\item[2) Success rate.] Targeted attack success rate, 
	$\mathrm{ASR}=\frac{\#(f(\boldsymbol{x}')=\boldsymbol{t})}{\#(\boldsymbol{x})}$, 
	is the proportion of adversarial images of fooling OCR models to output targeted text. 
	\item[3) Attack efficiency.] $\mathrm{I_{avg}}$ is average iterations of images to reach 100\% ASR.
\end{description}

 
%


\subsection{Letter-level Attack Performance}



Table~\ref{tab:adv_example} illustrates generated adversarial examples.  We find that the perturbations of basic attacks are quite obvious, especially when we want to replace the entire word, while the watermarks help hide the perturbations from human eyes.  

\begin{figure}[tb]
	\centering
	\includegraphics[width=\linewidth]{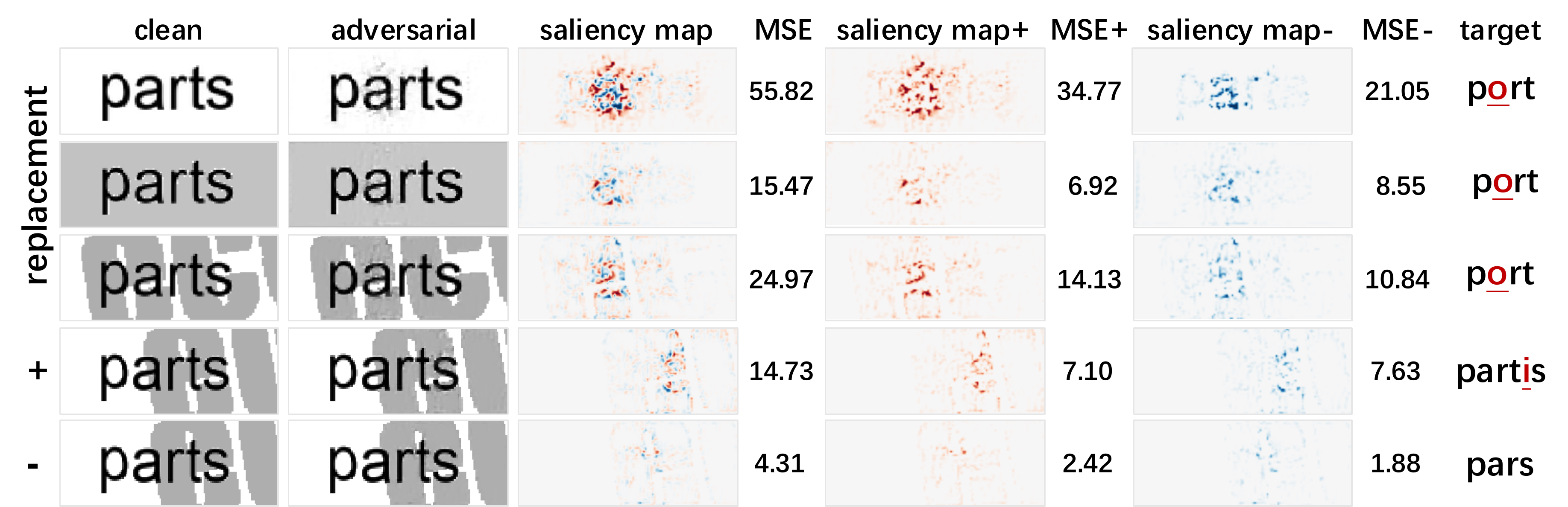}
	\caption{Saliency map visualization. First three lines are replacement. They have clean, gray and watermark backgrounds, respectively. Last two lines are insertion(+) and deletion(-). We fetch the positive part of saliency map to generate saliency map+ and the negative part to generate saliency map-. MSE, MSE+, MSE- represent the perturbation level in the saliency map, saliency map+ and saliency map-, respectively.}
	\label{fig:saliency_map}
\end{figure}

To intuitively analyze the effects of watermarks on our attacks, in Fig.~\ref{fig:saliency_map}, we first illustrate the saliency map that highlights the influence of each pixel on the target output $\boldsymbol{t}$.  
In replacement cases, we observe that the white background needs more perturbations, both positive (+) and negative (-), while the gray background requires 72\% less perturbations due to reduced contrast.  
Watermarks approximate the gray background and look more natural.  
Specifically, the attack of the white background also adds significant \texttt{a}-shaped negative perturbations to weaken the letter \texttt{a}, while the other two require negative perturbations less than 50\% due to lower contrast. 
In addition, the attacks add perturbations to neighboring letters, as the sequence-based OCR models also consider them. 

\begin{figure}[tb]
\centering
\subfigure[Grad-based attacks.]{
\begin{minipage}[]{0.26\linewidth}
\centering
\includegraphics[width=\linewidth,height=6cm]{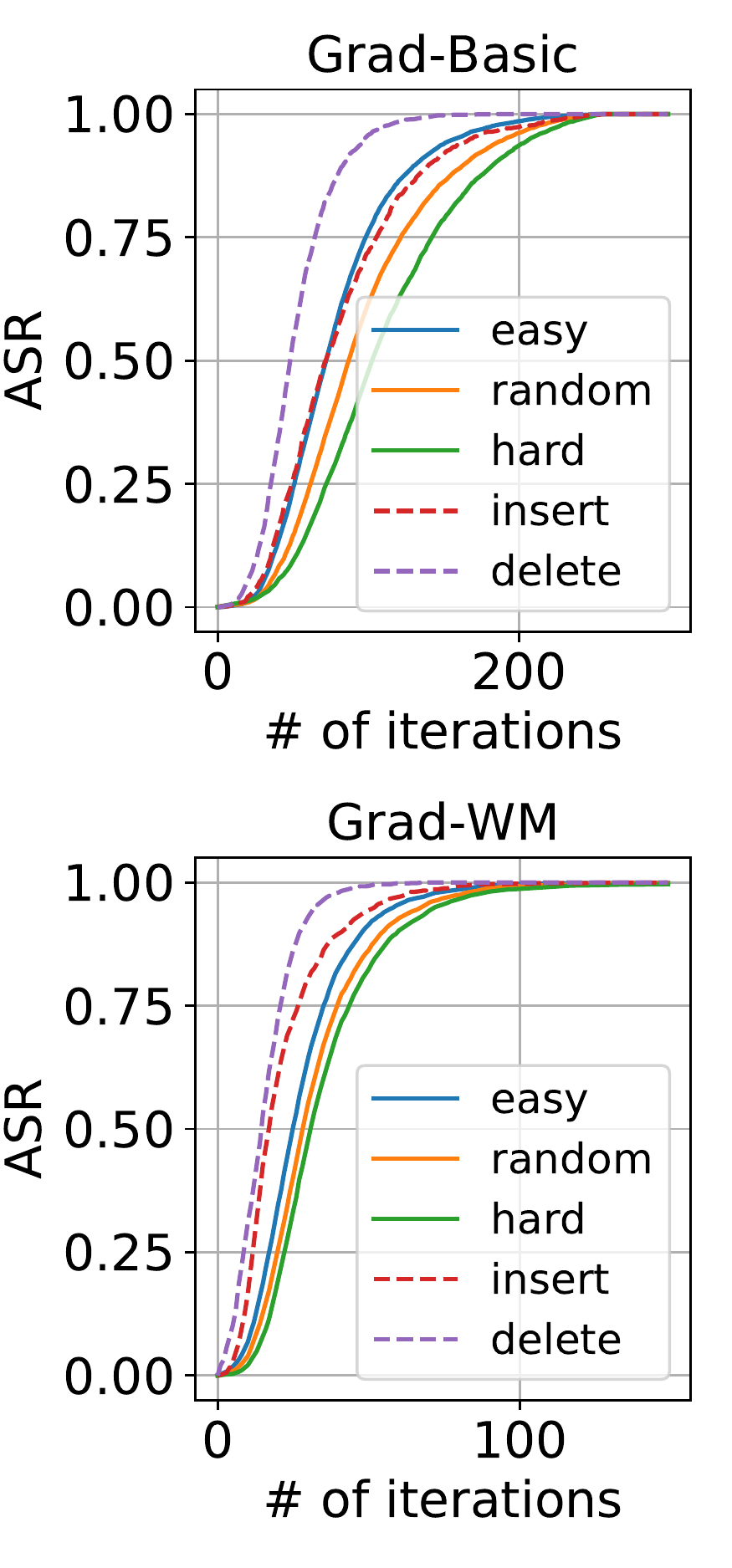}
\end{minipage}%
\label{fig:grad-attack}
}%
\subfigure[Opt-based attacks.]{
\begin{minipage}[]{0.26\linewidth}
\centering
\includegraphics[width=\linewidth,height=6cm]{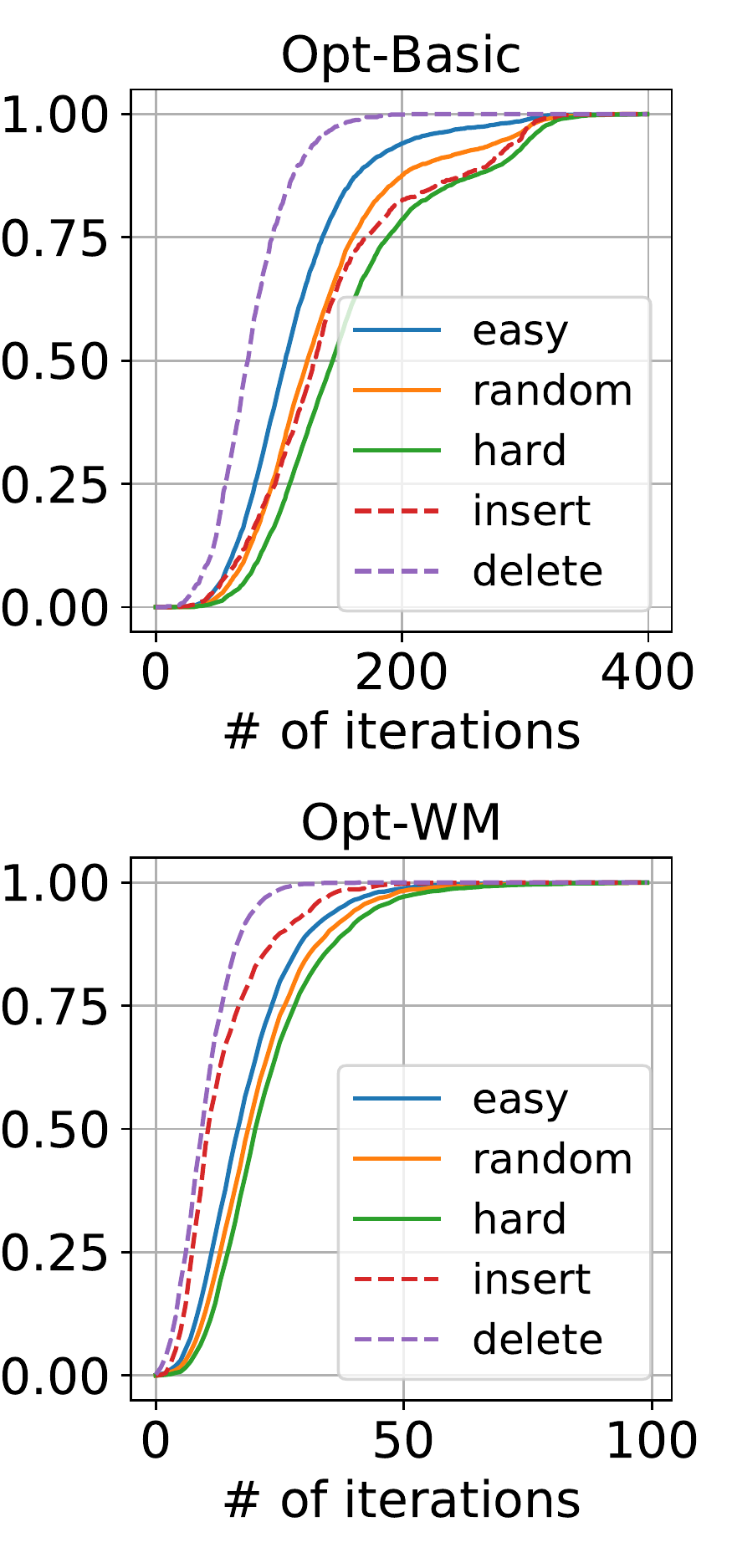}
\end{minipage}%
\label{fig:opt-attack}
}%
\subfigure[Other comparisions]{
\begin{minipage}[]{0.26\linewidth}
\centering
\includegraphics[width=\linewidth,height=6cm]{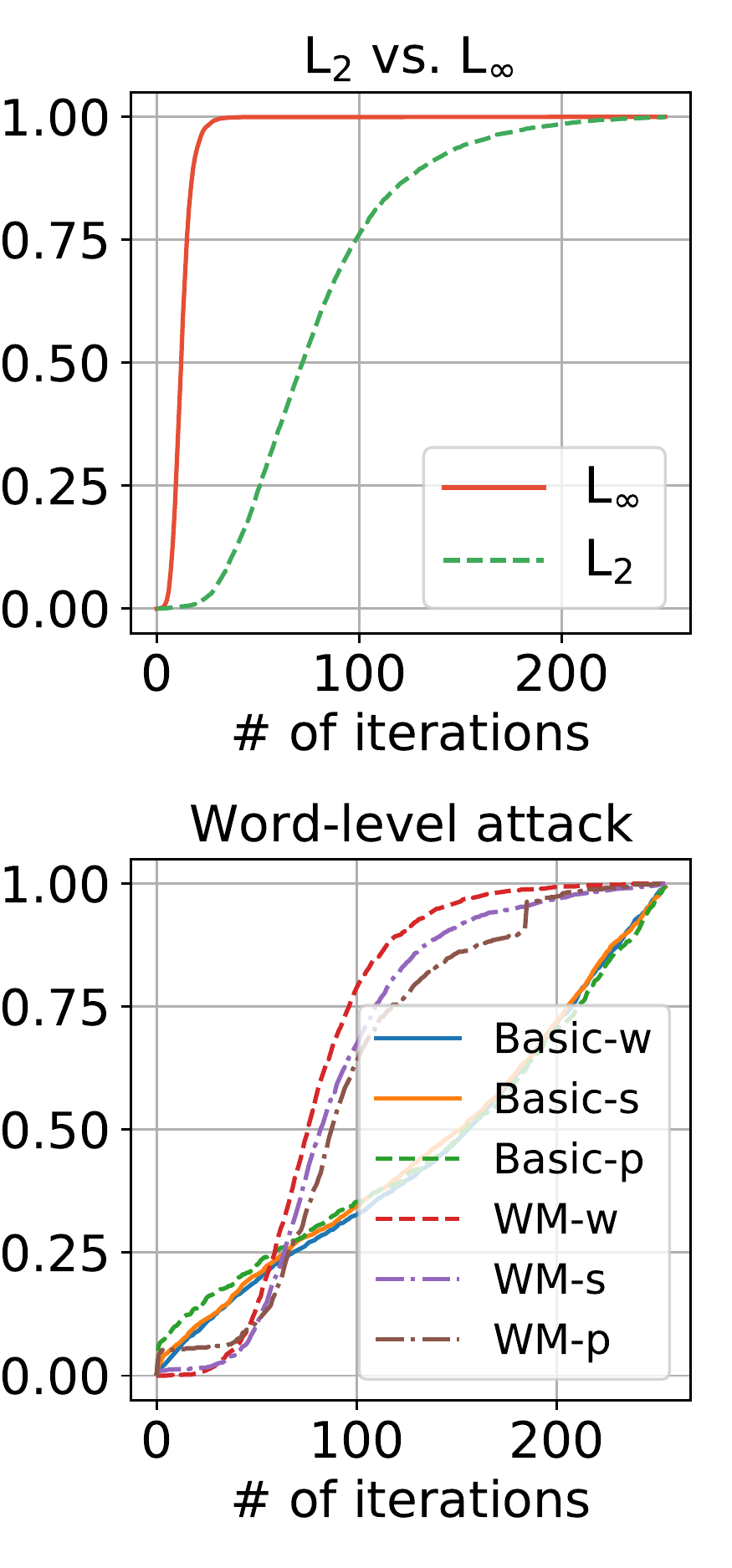}
\end{minipage}%
\label{fig:l2vslinf}
\label{fig:word-attack}
}%

\centering
\caption{Attack efficiency in word images with Arial font. }
\label{fig:attack_speed}
\end{figure}

Table~\ref{tab:difficulty} and Fig.~\ref{fig:attack_speed} show the quantitative analysis of attack performance and use different targets in letter-level attacks.
In Fig.~\ref{fig:attack_speed}, a sharper slope indicates a higher efficiency, that reaches a higher ASR in a fixed number of iterations.
ASRs of all attacks are 100\%.  
We observe that: 
1) MSEs of watermark attacks are only 40\% of basic attacks’ on average, confirming to the intuitive analysis above.  
2) Watermark attacks only require 78\% fewer $\mathrm{I_{avg}}$ on average, and have around 3 to 8 times sharper slope than basic attacks, showing the significant improvement of attack efficiency.  
3) Not surprisingly, hard cases require both more perturbations (higher MSE) and more iterations (larger $\mathrm{I_{avg}}$) with lower efficiency (flatter slope), due to lack of  similarity between two letters.  
4) In addition, deletions are easier than insertions and replacements. This is because OCR models are sensitive to perturbations, and classify fuzzy letters into blank tokens that will be ignored by CTC in the output. Intuitively, in the saliency map of Fig.~\ref{fig:saliency_map}, perturbations of deletion are much slighter than other cases. Thus few perturbations can achieve deletion.   
5) Courier font is easier to attack because it is thinner than other fonts.
Thus it requires less perturbations.  
6) Comparing gradient-based and optimization-based methods, the perturbation level is higher in gradient-based methods, no matter if they have watermarks or not. Because perturbations of optimization-based methods are not constrained by $\epsilon$.   



\begin{table}[tb]
\caption{Letter-level attacks using Grad-Basic, Grad-WM, Opt-Basic, Opt-WM attacks. Last line is the target output. We denote each font with their first letter.}
\label{tab:difficulty}
\resizebox{\textwidth}{!}{
\begin{tabular}{|c|c|cc|cc|cc|cc|cc|cc|cc|cc|cc|cc|}
\hline
\multicolumn{2}{|c|}{\multirow{4}{*}{}} &
  \multicolumn{10}{c|}{Gradient-based} &
  \multicolumn{10}{c|}{Optimization-based} \\ \cline{3-22}

\multicolumn{2}{|c|}{\multirow{3}{*}{}} &
  \multicolumn{6}{c|}{replacement} &
  \multicolumn{2}{c|}{\multirow{2}{*}{insertion}} &
  \multicolumn{2}{c|}{\multirow{2}{*}{deletion}} &
  \multicolumn{6}{c|}{replacement} &
  \multicolumn{2}{c|}{\multirow{2}{*}{insertion}} &
  \multicolumn{2}{c|}{\multirow{2}{*}{deletion}} \\ \cline{3-8} \cline{13-18}
  
\multicolumn{2}{|c|}{} &
  \multicolumn{2}{c|}{easy} &
  \multicolumn{2}{c|}{random} &
  \multicolumn{2}{c|}{hard} &
  \multicolumn{2}{c|}{} &
  \multicolumn{2}{c|}{} & 
   \multicolumn{2}{c|}{easy} &
  \multicolumn{2}{c|}{random} &
  \multicolumn{2}{c|}{hard} &
  \multicolumn{2}{c|}{} &
  \multicolumn{2}{c|}{}  \\ \cline{3-22} 
  
 \multicolumn{2}{|c|}{}  & 
\scriptsize{MSE} & \scriptsize{$\mathrm{I_{avg}}$} & 
\scriptsize{MSE} & \scriptsize{$\mathrm{I_{avg}}$} & 
\scriptsize{MSE} & \scriptsize{$\mathrm{I_{avg}}$} & 
\scriptsize{MSE} & \scriptsize{$\mathrm{I_{avg}}$} & 
\scriptsize{MSE} & \scriptsize{$\mathrm{I_{avg}}$} &
\scriptsize{MSE} & \scriptsize{$\mathrm{I_{avg}}$} & 
\scriptsize{MSE} & \scriptsize{$\mathrm{I_{avg}}$} & 
\scriptsize{MSE} & \scriptsize{$\mathrm{I_{avg}}$} & 
\scriptsize{MSE} & \scriptsize{$\mathrm{I_{avg}}$} & 
\scriptsize{MSE} & \scriptsize{$\mathrm{I_{avg}}$}   \\ \hline

\multirow{5}{*}{\rotatebox{90}{{Basic}} } 
 & \scriptsize{C} & 10.5 & 59 & 14.0 & 74 & 17.0 & 70 & 11.6 & 50 & 3.2 & 21 & 25.4 & 266 & 30.3 & 313 & 36.7 & 321 & 25.4 & 309 & 13.6 & 43  \\ 
 & \scriptsize{G} & 27.4 & 43 & 32.8 & 99 & 37.3 & 104 & 22.1 & 83 & 17.3 & 55 & 52.0 & 292 & 59.4 & 318 & 67.5 & 328 & 41.6 & 337 & 45.0 & 169 \\ 
 & \scriptsize{H} & 27.0 & 51 & 33.6 & 113 & 38.6 & 113 & 23.0 & 70 & 16.7 & 43 & 52.1 & 301 & 60.2 & 328 & 68.2 & 340 & 47.1 & 321 & 45.0 & 178 \\  
 & \scriptsize{T} & 26.4 & 62 & 31.5 & 85 & 35.8 & 109 & 20.3 & 98 & 17.2 & 68 & 49.9 & 294 & 56.1 & 324 & 61.6 & 345 & 41.7 & 314 & 44.3 & 172 \\ 
 & \scriptsize{A} & 29.8 & 51 & 36.7 & 73 & 42.5 & 66 & 24.3 & 88 & 19.2 & 59 & 56.3 & 304 & 65.3 & 327 & 73.8 & 341 & 48.3 & 324 & 51.0 & 176 \\  

 \hline
\multicolumn{2}{|c|}{} & \multicolumn{2}{c|}{\begin{minipage}[]{0.06\textwidth}
 \centerline{ \includegraphics[height=0.5cm]{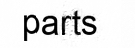}}
  \end{minipage}} & \multicolumn{2}{c|}{\begin{minipage}[]{0.06\textwidth}
 \centerline{ \includegraphics[height=0.5cm]{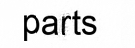}}
  \end{minipage}} &
 \multicolumn{2}{c|}{\begin{minipage}[]{0.1\textwidth}
 \centerline{ \includegraphics[height=0.5cm]{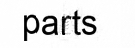}}
  \end{minipage}}  & \multicolumn{2}{c|}{\begin{minipage}[]{0.06\textwidth}
 \centerline{ \includegraphics[height=0.5cm]{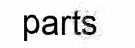}}
  \end{minipage}} & \multicolumn{2}{c|}{\begin{minipage}[]{0.06\textwidth}
 \centerline{ \includegraphics[height=0.5cm]{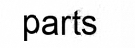}}
  \end{minipage}}  &
  \multicolumn{2}{c|}{\begin{minipage}[]{0.06\textwidth}
 \centerline{ \includegraphics[height=0.5cm]{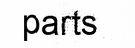}}
  \end{minipage}} & 
  \multicolumn{2}{c|}{\begin{minipage}[]{0.06\textwidth}
 \centerline{ \includegraphics[height=0.5cm]{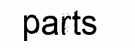}}
  \end{minipage}} &
 \multicolumn{2}{c|}{\begin{minipage}[]{0.06\textwidth}
 \centerline{ \includegraphics[height=0.5cm]{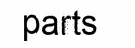}}
  \end{minipage}}  & 
  \multicolumn{2}{c|}{\begin{minipage}[]{0.06\textwidth}
 \centerline{ \includegraphics[height=0.5cm]{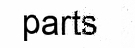}}
  \end{minipage}} & 
  \multicolumn{2}{c|}{\begin{minipage}[]{0.06\textwidth}
 \centerline{ \includegraphics[height=0.5cm]{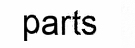}}
  \end{minipage}} \\  
\hline
\multirow{5}{*}{\rotatebox{90}{{WM}}}
 & \scriptsize{C} & 2.8 & 30 & 3.6 & 18 & 4.3 & 27 & 3.6 & 21 & 0.7 & 8 & 16.7 & 116 & 20.1 & 96 & 20.4 & 95 & 31.1 & 29 & 3.2 & 13 \\ 
 & \scriptsize{G} & 7.8 & 15 & 8.9 & 33 & 9.8 & 30 & 5.1 & 39 & 3.5 & 21 & 31.6 & 30 & 35.1 & 32 & 38.3 & 37 & 21.7 & 12 & 16.2 & 9 \\ 
 & \scriptsize{H} & 8.4 & 9 & 10.0 & 52 & 11.2 & 52 & 6.3 & 23 & 3.7 & 19 & 33.3 & 31 & 37.0 & 42 & 38.8 & 53 & 25.1 & 13 & 16.5 & 9 \\ 
 & \scriptsize{T} & 7.3 & 15 & 8.3 & 20 & 9.3 & 34 & 4.5 & 7 & 3.4 & 21 & 30.3 & 22 & 33.9 & 26 & 35.9 & 36 & 19.2 & 11 & 15.4 & 8 \\ 
 & \scriptsize{A} & 9.4 & 13 & 11.1 & 14 & 12.7 & 25 & 6.2 & 33 & 4.4 & 20 & 37.2 & 30 & 40.4 & 45 & 43.6 & 50 & 25.4 & 16 & 19.4 & 10 \\ 
  \hline
\multicolumn{2}{|c|}{} & 
\multicolumn{2}{c|}{\begin{minipage}[]{0.06\textwidth}
 \centerline{ \includegraphics[height=0.5cm]{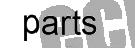}}
  \end{minipage}} & 
  \multicolumn{2}{c|}{\begin{minipage}[]{0.06\textwidth}
 \centerline{ \includegraphics[height=0.5cm]{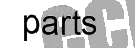}}
  \end{minipage}} &
 \multicolumn{2}{c|}{\begin{minipage}[]{0.06\textwidth}
 \centerline{ \includegraphics[height=0.5cm]{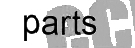}}
  \end{minipage}}  & 
  \multicolumn{2}{c|}{\begin{minipage}[]{0.06\textwidth}
 \centerline{ \includegraphics[height=0.5cm]{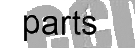}}
  \end{minipage}} & 
  \multicolumn{2}{c|}{\begin{minipage}[]{0.06\textwidth}
 \centerline{ \includegraphics[height=0.5cm]{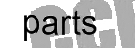}}
  \end{minipage}}  & \multicolumn{2}{c|}{\begin{minipage}[]{0.06\textwidth}
 \centerline{ \includegraphics[height=0.5cm]{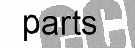}}
  \end{minipage}} & 
  \multicolumn{2}{c|}{\begin{minipage}[]{0.06\textwidth}
 \centerline{ \includegraphics[height=0.5cm]{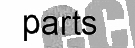}}
  \end{minipage}} &
 \multicolumn{2}{c|}{\begin{minipage}[]{0.06\textwidth}
 \centerline{ \includegraphics[height=0.5cm]{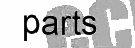}}
  \end{minipage}}  & 
  \multicolumn{2}{c|}{\begin{minipage}[]{0.06\textwidth}
 \centerline{ \includegraphics[height=0.5cm]{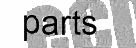}}
  \end{minipage}} & 
  \multicolumn{2}{c|}{\begin{minipage}[]{0.06\textwidth}
 \centerline{ \includegraphics[height=0.5cm]{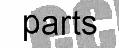}}
  \end{minipage}} \\  
  \hline
\multicolumn{2}{|c|}{output} & \multicolumn{2}{c|}{pants} & \multicolumn{2}{c|}{pacts} & \multicolumn{2}{c|}{pasts}  & \multicolumn{2}{c|}{partis} & \multicolumn{2}{c|}{pars} & \multicolumn{2}{c|}{pants} & \multicolumn{2}{c|}{pacts} & \multicolumn{2}{c|}{pasts}  & \multicolumn{2}{c|}{partis} & \multicolumn{2}{c|}{pars} \\  
\hline
\end{tabular}}
\end{table}

\subsection{Word-Level Attack Performance}  

\begin{table}[tb]
	\centering
	\caption{Word-level attacks in word, sentence and paragraph images.}
	\label{tab:word_attack}
\resizebox{\textwidth}{!}{
	\begin{tabular}{|c|c|ccc|c|ccc|c|ccc|c|}
		\hline
		\multicolumn{2}{|c|}{\multirow{2}{*}{}} & 
		\multicolumn{4}{c|}{word image} & 
		\multicolumn{4}{c|}{ sentence image } &
		\multicolumn{4}{c|}{ paragraph image } \\ \cline{3-14} 
		
		\multicolumn{2}{|c|}{} & 
		MSE & PSNR & SSIM & $\mathrm{I_{avg}}$ & 
		MSE & PSNR & SSIM & $\mathrm{I_{avg}}$ &
		MSE & PSNR & SSIM & $\mathrm{I_{avg}}$ \\ \hline
		
		\multirow{6}{*}{\makecell{Grad-\\ Basic}}
		& Courier & 50.6 & 31.1 & 0.944 & 235 & 5.8 & 40.5 & 0.993 & 153 & 7.0 & 39.7 & 0.993 & 113 \\ 
		& Georgia & 118.6 & 27.4 & 0.900 & 326 & 14.6 & 36.5 & 0.988 & 239 & 21.3 & 34.9 & 0.986 & 203 \\ 
		& Helvetica & 124.0 & 27.2 & 0.894 & 254 & 14.2 & 36.6 & 0.988 & 238 & 22.5 & 34.6 & 0.984 & 233 \\ 
		& Times & 114.4 & 27.5 & 0.904 & 291 & 13.2 & 36.9 & 0.989 & 201 & 17.4 & 35.7 & 0.989 & 164 \\ 
		& Arial & 133.9 & 26.9 & 0.888 & 222 & 15.8 & 36.1 & 0.987 & 273 & 23.2 & 34.5 & 0.984 & 242 \\ 
		\cline{2-14}
		
		& { example }  & 
		\multicolumn{4}{c|}{\begin{minipage}[]{0.09\textwidth}
				\centerline{ \includegraphics[height=0.5cm]{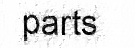}}
		\end{minipage}}  &
		\multicolumn{8}{c|}{\begin{minipage}[]{0.09\textwidth}
				\centerline{ \includegraphics[height=0.5cm]{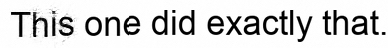}}
		\end{minipage}} \\  \hline
		\multirow{5}{*}{\makecell{Grad\\-WM}} 
		& Courier & 12.9 & 37.0 & 0.993 & 51 & 3.4 & 42.8 & 0.999 & 60 & 8.7 & 38.7 & 0.998 & 57 \\ 
		& Georgia & 35.4 & 32.6 & 0.985 & 131 & 5.2 & 41.0 & 0.999 & 90 & 9.2 & 38.5 & 0.998 & 81 \\ 
		& Helvetica & 40.0 & 32.1 & 0.984 & 124 & 5.4 & 40.8 & 0.999 & 91 & 11.8 & 37.4 & 0.998 & 96 \\ 
		& Times & 34.8 & 32.7 & 0.985 & 160 & 4.5 & 41.6 & 0.999 & 79 & 6.7 & 39.9 & 0.999 & 74 \\ 
		& Arial & 44.6 & 31.6 & 0.982 & 138 & 6.1 & 40.3 & 0.999 & 98 & 12.0 & 37.3 & 0.998 & 98 \\ 
		\cline{2-14}
		
		& { example}  & 
		\multicolumn{4}{c|}{\begin{minipage}[]{0.09\textwidth}
				\centerline{ \includegraphics[height=0.5cm]{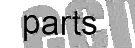}}
		\end{minipage}}  &
		\multicolumn{8}{c|}{\begin{minipage}[]{0.09\textwidth}
				\centerline{ \includegraphics[height=0.5cm]{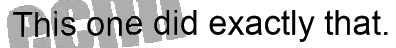}}
		\end{minipage}} \\  \hline
		
		\multicolumn{2}{|c|}{ \makecell{target output}}  & 
		\multicolumn{4}{c|}{taupe}  &
		\multicolumn{8}{c|}{Tale one did exactly that.}  \\  \hline
	\end{tabular}}
\end{table}

We can still achieve 100\% ASR in the word-level attacks, but both MSE and $\mathrm{I_{avg}}$ of word-level attacks are significantly higher than those of letter-level attacks.   

We perform word-level attacks in word, sentence and paragraph images. 
Table~\ref{tab:word_attack} and Fig.~\ref{fig:l2vslinf} summarize results in word-level attacks. 
Due to limited space, we only show results of gradient-based methods. The optimization-based methods have similar insights. 
In word-level attacks, we have similar observations as letter-level attacks. 
1) Watermark attacks spend 50\% less $\mathrm{I_{avg}}$ with higher efficiency (sharper slope) generating adversarial images than basic attacks.  
2) We observe 56\% lower MSE with watermarks averagely.  
3) It is easier to attack Courier font than other fonts for less MSE and $\mathrm{I_{avg}}$.  
4) Due to a larger number of pixels in sentence and paragraph images, their MSEs are much lower than MSE of word images, but the absolute number of affected pixels stays the same.  





\subsection{Effects of Hyper-parameter Settings }
\label{eval:parameters}
To better understand these attacks, we evaluate the effects of hyper-parameter settings, such as tradeoff $c$, $\mathrm{L_p}$-norms, font-weight, position of watermarks, etc.  

\para{Effects of plain watermarks.} 
First, we evaluate the effects of plain watermarks, i.e., initial watermarks without perturbations.  The accuracy of Courier, Georgia, Helvetica, Times, Arial drop to 0.066, 0.768, 0.531, 0.753 and 0.715, respectively, showing that it is quite trivial to launch \emph{untargeted attacks} on OCR. Also, it further confirms that thinner font, Courier, is highly sensitive to plain watermarks. In samples of incorrect recognition, we count the percentage of output text shorter than the ground-truth text, Courier(0.65), Georgia(0.12), Helvetica(0.08), Times(0.09), Arial(0.14). It's easy to induce that thinner font is more sensitive to perturbations resulting in losing letters like the deletion case.

\begin{figure}[tb]
	\centering
	\includegraphics[width=\linewidth]{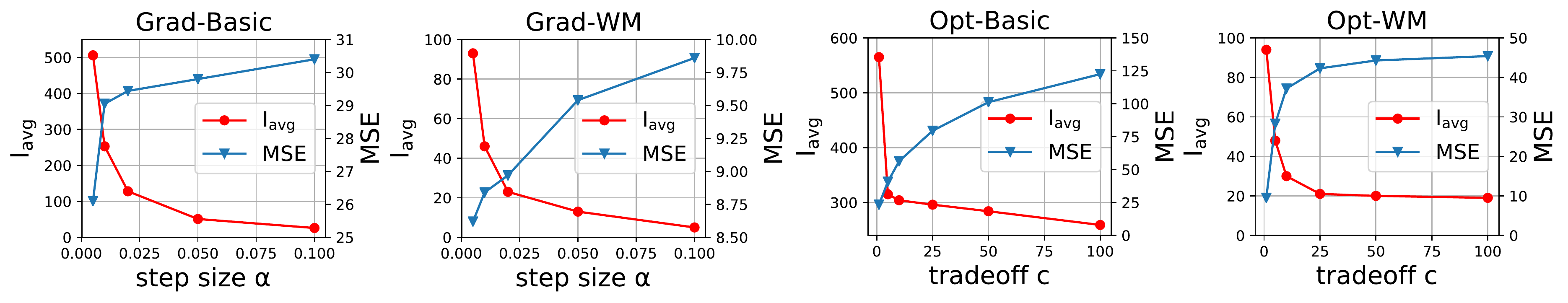}
	\caption{In easy case of Arial word images, Grad-Basic and Grad-WM with different step size $\alpha=0.005, 0.010, 0.020, 0.050, 0.100$. In easy case of Arial word images, Opt-Basic and Opt-WM  with different tradeoff $c= 1, 5, 10, 25, 50, 100$.}
	\label{fig:tradeoff}
\end{figure}

\para{Tradeoff between efficiency and visual quality. }
In both gradient-based and optimization-based methods, we need to set tradeoff parameters to balance between efficiency ($\mathrm{I_{avg}}$) and visual quality (MSE).
In Fig.~\ref{fig:tradeoff}, we plot the change of $\mathrm{I_{avg}}$ and MSE along with distinct tradeoff settings.
In gradient-based methods, we can reduce perturbation level with smaller step size $\alpha$, at the cost of attack efficiency (raised $\mathrm{I_{avg}}$).
In optimization-based methods, the main quality-efficiency tradeoff parameter $c$ balances the targeted loss and the $\mathrm{L_p}$-norm distance. 
We find that a smaller $c$ makes the optimized objective function pay more attention to reduce the perturbation level with larger $\mathrm{I_{avg}}$. 
Thus we can use the binary search for $c$ to find adversarial examples with a lower perturbation level under the premise of maintaining perfect ASR. 
In addition, in various hyperparameter settings, the performance of watermark attacks is better than the basic attacks.

\para{Setting $\mathrm{L_{p}}$-norms.}
\label{sec:lp-norm}
\textit{${L_\infty}$-norm} for measuring the maximum variation of perturbations has the same perturbed value at each perturbed pixel, and thus significantly narrows down operation space. Even more, observing $\mathrm{L_\infty}$-norm examples in Table~\ref{tab:lp-norm}, $\mathrm{L_\infty}$-norm causes severe background noise. 
\textit{${L_2}$-norm} yields perturbed values varying in perturbed pixels, and thus offers better flexibility to perform stronger adversarial attacks. 
Although $\mathrm{L_2}$-norm is lower efficient than $\mathrm{L_{\infty}}$-norm in Fig.~\ref{fig:l2vslinf}, $\mathrm{L_2}$-norm's image quality metrics, MSE, PSNR and SSIM, in Table~\ref{tab:lp-norm},  all are better than those of $\mathrm{L_{\infty}}$-norm. 
Intuitively, $\mathrm{L_2}$-norm examples also have better visual quality. 
Therefore, we choose the $\mathrm{L_2}$-norm in our experiments.

\para{Bold fonts settings.} 
We also investigate the bold version of the five fonts. Table~\ref{tab:lp-norm} shows most bold fonts require slightly more MSE and $\mathrm{I_{avg}}$ to attack successfully, except for Courier requiring doubling MSE. This is not surprising, as bold fonts contain more useful pixels per letter. So they need more perturbations.

\begin{table}[tb]
\centering
\caption{Comparision of $\mathrm{L_2}$-norm and $\mathrm{L_{\infty}}$-norm in Grad-Basic. Examples' target output are ``p\underline{\textbf{o}}rts". Fonts are denoted by their initial letters. Bold fonts marked as 'b'.}
\label{tab:lp-norm}
\setlength{\tabcolsep}{0.1mm}{
\begin{tabular}{|c|ccc|c| c |c|ccc|c| c |ccc|c| c |}
\hline
\multirow{2}{*}{} & 
\multicolumn{5}{c|}{$\mathrm{L_2}$} & \multirow{2}{*}{} & \multicolumn{5}{c|}{$\mathrm{L_2}$} &
\multicolumn{5}{c|}{$\mathrm{L_{\infty}}$ } \\ \cline{2-6} \cline{8-17}
 & \scriptsize{MSE} & \scriptsize{PSNR} & \scriptsize{SSIM} & \scriptsize{$\mathrm{I_{avg}}$} &  \scriptsize{example} &
 &  \scriptsize{MSE} & \scriptsize{PSNR} & \scriptsize{SSIM} & \scriptsize{$\mathrm{I_{avg}}$} &  \scriptsize{example} &
     \scriptsize{MSE} & \scriptsize{PSNR} & \scriptsize{SSIM} & \scriptsize{$\mathrm{I_{avg}}$} &  \scriptsize{example}  \\ 
\hline
Cb & 23.6 & 34.4 & 0.977 & 69 & \begin{minipage}[]{0.09\textwidth}
 { \includegraphics[height=0.4cm]{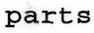}}
  \end{minipage} & C & 10.5 & 37.9 & 0.988 & 59 & \begin{minipage}{0.09\textwidth}
 { \includegraphics[height=0.4cm]{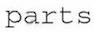}}
  \end{minipage} & 72.6 & 29.5 & 0.771 & 11 & \begin{minipage}{0.09\textwidth}
 { \includegraphics[height=0.4cm]{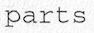}}
  \end{minipage} \\ 
Gb & 25.9 & 34.0 & 0.975 & 59 & \begin{minipage}[]{0.09\textwidth}
 { \includegraphics[height=0.5cm]{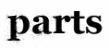}}
  \end{minipage} &  G & 27.4 & 33.7 & 0.974 & 43 & \begin{minipage}{0.09\textwidth}
 { \includegraphics[height=0.5cm]{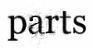}}
  \end{minipage} & 163.6 & 26 & 0.645 & 11 & \begin{minipage}{0.09\textwidth}
 { \includegraphics[height=0.5cm]{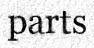}}
  \end{minipage} \\ 
Hb & 32.8 & 33.0 & 0.970 & 75 & \begin{minipage}[]{0.09\textwidth}
 { \includegraphics[height=0.5cm]{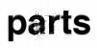}}
  \end{minipage} & H & 27.0 & 33.8 & 0.975 & 51 & \begin{minipage}{0.09\textwidth}
 { \includegraphics[height=0.5cm]{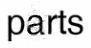}}
  \end{minipage} & 159.4 & 26.1 & 0.658 & 10 & \begin{minipage}{0.09\textwidth}
 { \includegraphics[height=0.5cm]{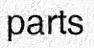}}
  \end{minipage} \\ 
Tb & 26.3 & 33.9 & 0.975 & 66 & \begin{minipage}[]{0.09\textwidth}
 { \includegraphics[height=0.5cm]{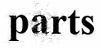}}
  \end{minipage} & T & 26.4 & 33.9 & 0.975 & 62 & \begin{minipage}{0.09\textwidth}
 { \includegraphics[height=0.5cm]{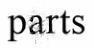}}
  \end{minipage} & 156.6 & 26.2 & 0.653 & 11 & \begin{minipage}{0.09\textwidth}
 { \includegraphics[height=0.5cm]{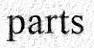}}
  \end{minipage} \\ 
Ab & 34.2 & 32.8 & 0.968 & 69 & \begin{minipage}[]{0.09\textwidth}
 { \includegraphics[height=0.5cm]{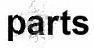}}
  \end{minipage} & A & 29.8 & 33.4 & 0.972 & 51 & \begin{minipage}{0.09\textwidth}
 { \includegraphics[height=0.5cm]{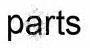}}
  \end{minipage} & 169.3 & 25.8 & 0.656 & 11 & \begin{minipage}{0.09\textwidth}
 { \includegraphics[height=0.5cm]{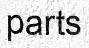}}
  \end{minipage} \\  
\hline

\end{tabular}}
\end{table}

\para{Attacking sequence-based OCRs.}
As OCR recognizes entire sequences rather than individual characters, we demonstrate that we can replace a letter even the watermark does not overlap with the letter, indicating we will not add perturbations over it. 
In this experiment, we shift the watermark about 10 pixels right to the target letter when generating initial watermarked images. 
Table~\ref{tab:long-attack} summarizes results that the attacks require 18 times MSE and  34 times $\mathrm{I_{avg}}$, and ASRs drop to less than 50\% except for the simplest Courier case. 
It confirms the fact that in sequential-based OCRs, the influence around the letter is not as important as that overlaying the letter.
Also, it reveals the necessity of finding the position of watermarks accurately to perform strong adversarial attacks.

\begin{table}[tb]
\noindent
\begin{minipage}{0.49\textwidth}
\centering
\caption{Non-overlapping Grad-WM in easy case of word images. Add a shift about 10 pixels right to the target letter.}
\label{tab:long-attack}
\begin{tabular}{|c|c|c|c|c|c| c |}
\hline
 & MSE  
 & $\mathrm{I_{avg}}$ 
 & ASR 
 & example 
 & output \\ \hline
Courier & 48.5 & 186 & 0.83 &  \begin{minipage}[]{0.2\textwidth}
 \centerline{ \includegraphics[height=0.45cm]{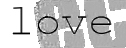}}
  \end{minipage} & \underline{\textbf{h}}ove\\ 
Georgia & 156.6 & 546 & 0.36 &  \begin{minipage}[]{0.2\textwidth}
 \centerline{ \includegraphics[height=0.45cm]{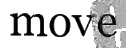}}
  \end{minipage} & mo\underline{\textbf{v}}e\\ 
Helvetica & 146.8 & 474  &0.45 &  \begin{minipage}[]{0.2\textwidth}
 \centerline{ \includegraphics[height=0.45cm]{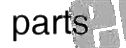}}
  \end{minipage} & pa\underline{\textbf{n}}ts \\ 
Times & 145.9  & 515 & 0.43  & \begin{minipage}[]{0.2\textwidth}
 \centerline{ \includegraphics[height=0.45cm]{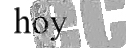}}
  \end{minipage} & \underline{\textbf{b}}oy \\ 
Arial & 153.5  & 463 & 0.44   & \begin{minipage}[]{0.2\textwidth}
 \centerline{ \includegraphics[height=0.45cm]{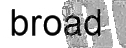}}
  \end{minipage} & br\underline{\textbf{e}}ad \\ 
\hline
\end{tabular}
\end{minipage}
\hspace{0.1cm}
\begin{minipage}{0.49\textwidth}
\centering
\caption{Protection mechanism. Acc and Acc* are the prediction accuracy with and without protection, respectively. }
\label{tab:protection}
\begin{tabular}{|c|c|c|cc|c|}
\hline
 & MSE & $\mathrm{I_{avg}}$ & Acc & Acc* & example \\ \hline
 
Courier & 0.6 & 5 & 1.0 & 0.066 & \begin{minipage}[]{0.2\textwidth}
 \centerline{ \includegraphics[height=0.45cm]{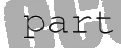}}
  \end{minipage} \\ 
Georgia & 0.2 & 1 & 1.0 & 0.768 & \begin{minipage}[]{0.2\textwidth}
 \centerline{ \includegraphics[height=0.45cm]{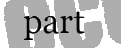}}
  \end{minipage}\\ 
Helvetica & 0.2 & 1 & 1.0 & 0.531 & \begin{minipage}[]{0.2\textwidth}
 \centerline{ \includegraphics[height=0.45cm]{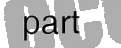}}
  \end{minipage}\\ 
Times & 0.2 & 1 & 1.0 & 0.753 & \begin{minipage}[]{0.2\textwidth}
 \centerline{ \includegraphics[height=0.45cm]{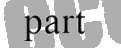}}
  \end{minipage}\\  
Arial & 0.2  & 1 & 1.0 & 0.715 & \begin{minipage}[]{0.2\textwidth}
 \centerline{ \includegraphics[height=0.45cm]{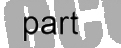}}
  \end{minipage}\\ 
\hline 
\end{tabular}
\end{minipage}%
\end{table}

\subsection{Extensions and Applications of Watermark Attacks}

\begin{figure}[tb]
	\centering
	\includegraphics[width=\linewidth]{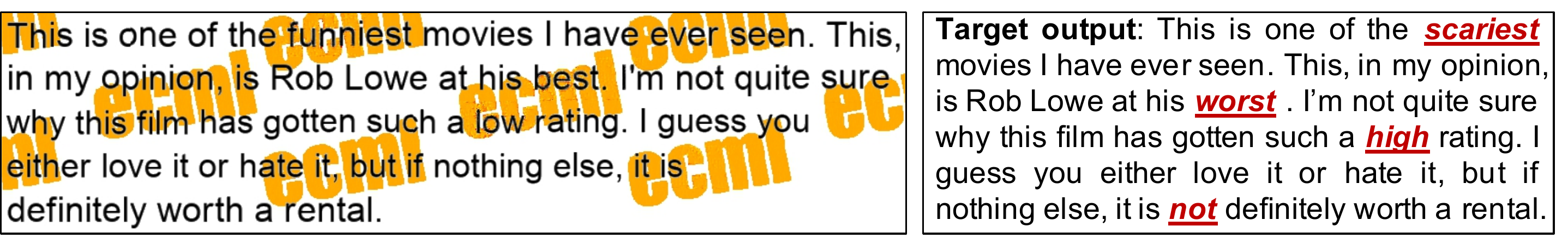}
	\caption{Full-color watermarks on a paragraph image. \scriptsize{MSE/PSNR/SSIM: 9.36/38.42/0.998}.}
	\label{fig:color_wm}
\end{figure}

\para{Full-color watermarks.}
Sometimes adding gray watermarks still hurts human readability.  We use the fact that modern OCR first transforms colored images into gray ones before recognition. Colored watermarks significantly improve overall readability when mixed with black texts. 
Fig.~\ref{fig:color_wm} shows a colored-watermark example of altering a positive movie review into a negative one by replacing and inserting words in it.  
Note that not all watermarks are malicious (e.g., the watermark on ``I have ever seen'' of Fig.~\ref{fig:color_wm} does not include adversarial perturbations).  We evenly distribute watermarks over the paragraph image to make it look more similar to a naturally watermarked paragraph image in the real-world scenario.  

\begin{figure}[tb]
\centering
\includegraphics[width=\linewidth]{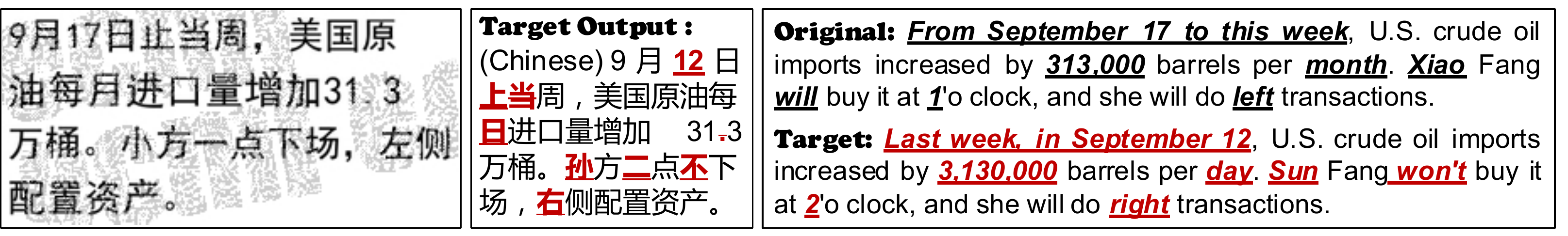}
\caption{A Chinese paragraph example. \scriptsize{MSE/PSNR/SSIM: 735.34/19.46/0.697}.}
\label{fig:chinese_sample}
\end{figure}


\para{Attacking Chinese Characters.}
In addition to English, we show that the method is applicable to other languages. Fig.~\ref{fig:chinese_sample} shows a Chinese example where we almost altered all important information, and its perturbation level is much larger than that in Fig.~\ref{fig:color_wm}, because of the complex structure of Chinese characters.

\para{Using FAWA to enhance the OCR readability of watermarked contents.}  
Sometimes we want people to notice vital watermarks (e.g., urgent, confidential), but we do not want them to affect OCR's accuracy. 
In such case, we generate accuracy-enhancing watermarks by setting the ground-truth as the target. 
In Table~\ref{tab:protection}, with few $\mathrm{I_{avg}}$ and MSEs, Acc (accuracy with protection mechanism) increases to 1.0 compared with the low Acc* (accuracy of initial watermarked images). 
It works because protective perturbations strengthen the features of the ground-truth and boost the confidence of the original, making it more ``similar'' to the target (i.e., the ground-truth). As the target is the same as the original, both MSE and $\mathrm{I_{avg}}$ stay low. 
So we can use FAWA as a protection mechanism to produce both  human-friendly images and OCR-friendly images.

\section{Conclusion}

DNN-based OCR systems are vulnerable to adversarial examples.  
On the text images, we hide perturbations in watermarks, making adversarial examples look natural to human eyes. 
Specifically, we develop FAWA that automatically generate the adversarial watermarks targeting sequence-based OCR models. 
In extensive experiments, while maintaining perfect attack success rate, FAWA exhibits the outstanding attack capability. For example, FAWA boosts the attack speed up to 8 times, and reduces the perturbation level lower than 40\% on average. In word, sentence and paragraph contexts, FAWA works well with letter-level and word-level targets.
We further extend our natural watermarked adversarial examples in many scenarios, such as full-color watermarks to increase the text readability, applicability for other languages, protection mechanism for enhancing OCR's accuracy.
We believe human-eye-friendly adversarial samples are applicable in many other scenarios, and we plan to explore them as future work.

\section*{Acknowledgments}
This work is supported in part by the National Natural Science Foundation of China (NSFC) Grant 61532001 and the Zhongguancun Haihua Institute for Frontier Information Technology.

 \bibliographystyle{splncs04}
 \bibliography{main.bib}

\end{document}